%% file: acl_latex.tex
\pdfoutput=1

\documentclass[11pt]{article}

\usepackage[]{acl}

\usepackage{times}
\usepackage{latexsym}
\usepackage{graphicx}
\usepackage{hhline}
\usepackage{amsmath}
\usepackage{amssymb}

\usepackage[T1]{fontenc}

\usepackage[utf8]{inputenc}

\usepackage{microtype}

\newcommand*\samethanks[1][\value{footnote}]{\footnotemark[#1]}

\newcommand{\abovezeroshot}{>ZS (\%)}
\newcommand{\aboverandom}{>$\E_{P_r}$ (\%)}

\DeclareMathOperator{\argmax}{argmax}
\DeclareMathOperator{\E}{\mathbb{E}}
\newcommand{\Tau}{\mathcal{T}}
\DeclareMathOperator{\Eval}{Eval}

\title{Reducing Retraining by Recycling Parameter-Efficient Prompts}

\author{Brian Lester\thanks{\hspace{0.5em}Equal Contribution} \enskip Joshua Yurtsever\samethanks \enskip Siamak Shakeri \enskip Noah Constant \\
  Google Research \\
  \texttt{\{brianlester, jyurtsever, siamaks, nconstant\}@google.com} \\}

\begin{document}
\maketitle

\input{sections/abstract}
\input{sections/intro}
\input{sections/related_work}
\input{sections/methods}
\input{sections/results}
\input{sections/conclusion}
\input{sections/acknowledgements}

\bibliography{anthology,custom}

\appendix
\input{sections/appendix}

\end{document}

%% file: sections/abstract.tex
\begin{abstract}
Parameter-efficient methods are able to use a single frozen pre-trained large language model (LLM) to perform many tasks by learning task-specific soft prompts that modulate model behavior when concatenated to the input text. However, these learned prompts are tightly coupled to a given frozen model---if the model is updated, corresponding new prompts need to be obtained. In this work, we propose and investigate several approaches to ``Prompt Recycling'', where a prompt trained on a \textit{source model} is transformed to work with the new \textit{target model}. Our methods do not rely on supervised pairs of prompts, task-specific data, or training updates with the target model, which would be just as costly as re-tuning prompts with the target model from scratch. We show that recycling between models is possible (our best settings are able to successfully recycle $88.9\%$ of prompts, producing a prompt that out-performs baselines), but significant performance headroom remains, requiring improved recycling techniques.

\end{abstract}

%% file: sections/intro.tex
\section{Introduction}

Fine-tuning pre-trained large language models (LLMs) is the current de-facto approach for achieving state-of-the-art results in NLP \cite{radfordImprovingLanguageUnderstanding,devlinBERTPretrainingDeep2019,raffelExploringLimitsTransfer2020}. While the exact details of the strongest models shift over time, there has been a clear trend that bigger models have better performance \cite{kaplanScalingLawsNeural2020,raeScalingLanguageModels2022,chowdheryPaLMScalingLanguage2022}. As these models have grown, the computational resources required and engineering complexity have grown as well. This trend is especially challenging in the multi-task setting where each task creates a new fork of the model.

Several parameter-efficient methods have been developed to mitigate these problems. Some methods only train part of the model \cite{houlsbyParameterEfficientTransferLearning2019}, while others use specially crafted inputs---either discrete text tokens \cite{brownLanguageModelsAre2020} or learned soft prompts \cite{qinLearningHowAsk2021,zhongFactualProbingMASK2021,liPrefixTuningOptimizingContinuous2021,lesterPowerScaleParameter2021}. By swapping out small components tailored to individual tasks, one can have a multi-task model without the need to serve or store many copies.

The specific LLMs that serve as the best starting point for building downstream models change over time, due to several unavoidable factors. New facts need to be learned based on world events, word meanings drift \cite{kulkarniStatisticallySignificantDetection2015}, and new terms (e.g., ``COVID'') enter common usage, requiring periodic refreshes. Additionally, new research regularly yields improvements through changes to model architecture, size, and details of training. When a new pre-trained model is released, new versions of task-specific models created via fine-tuning need to be trained to take advantage of the improvements.

\begin{figure}[t!]
    \centering
    \includegraphics[width=0.95\columnwidth]{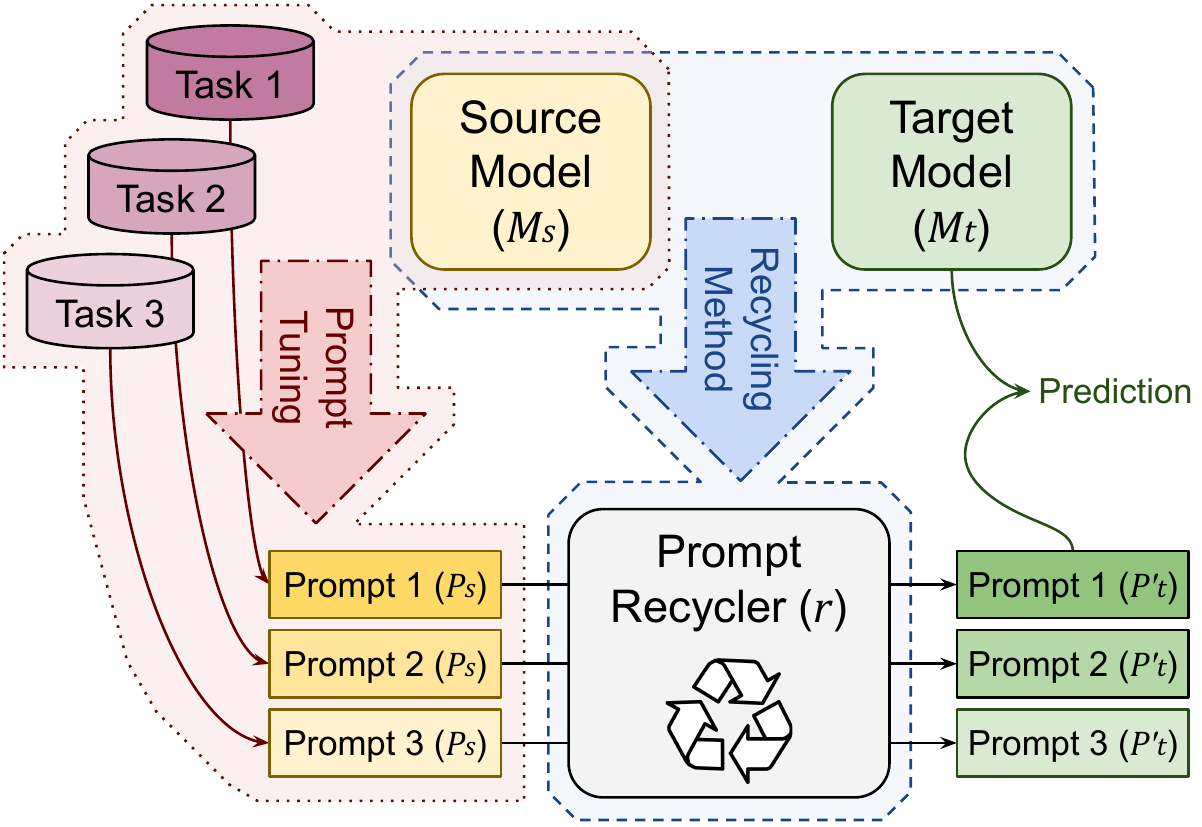} 
    \caption{We explore methods of building a task-agnostic Prompt Recycler ($r$) capable of transforming prompts ($P_s$) tuned for a source model ($M_s$) into prompts ($P'_t$) that can be used with a target model ($M_t$).}
    \label{fig:recycling}
\end{figure}

Just like fine-tuning, parameter-efficient methods are closely tied to the frozen model they were trained with. These approaches rely on making modifications to the activations of the model when presented with new input, to steer the model towards solving a new task. When the frozen model changes, the exact patterns of how it processes a given input change as well, and the modifications these approaches induce no longer make sense. Thus, parameter-efficient methods also require re-training when the frozen models change.

To mitigate this issue, we propose ``Prompt Recycling'' where soft prompts, learned via Prompt Tuning \cite{lesterPowerScaleParameter2021}, for a \textit{source model} $M_s$, are transformed to work with a \textit{target model} $M_t$, without requiring any training of the target model. This process is illustrated in Figure~\ref{fig:recycling}.

We present several recycling methods that center around the known correspondence between the embedded representation of a given word in each model. Our methods do not require paired prompts (two prompts that solve the same task, one learned using $M_s$ and one learned with $M_t$) or additional training of the prompt with $M_t$. Requiring either of these would likely result in methods that are no more computationally efficient than simply training a new prompt from scratch.

We demonstrate that our recycling methods can facilitate the transfer of prompts between models. Our best settings are able to successfully recycle $88.9\%$ of prompts, where we count success as producing a prompt that performs better than the target model's zero-shot capabilities. However, recycled performance still lags behind prompts trained with the target model from scratch (see Table~\ref{tab:perf}). Additionally we find that recycling from a larger source model to a smaller target model increases reliability and performance of the recycled prompt.

%% file: sections/related_work.tex
\section{Related Work}

Our vocab-to-vocab transformations are similar to cross-lingual embedding mapping \cite{mikolovExploitingSimilaritiesLanguages2013}, except that our mapping is applied to models trained on the same language.

The ``soft prompt transfer'' technique of \citet{vuSPoTBetterFrozen2022} is similar to ours in that prompts are reused between different models; however, \citet{vuSPoTBetterFrozen2022} focuses on transfer between different tasks with the same pre-trained model while we focus on transfer between different models trained with the same task. Additionally, their method assumes the prompt will be updated with training data for the new task. In contrast, our method does not update the prompt after recycling as that as it would remove the computational advantage over training a prompt on $M_t$ directly. The finding from \citet{vuSPoTBetterFrozen2022} that one prompt can be successfully used to initialize another suggests that using a recycled prompt as initialization could also be worth investigating.

Work from \citet{suTransferabilityPromptTuning2022} also investigates the transfer of prompts across models. However, their work focuses on knowledge transfer between very different pre-trained language models (different datasets, different architectures, etc.\@) with the aim of increasing final model performance, whereas our work approaches prompt recycling as a way to avoid re-training prompts when a frozen model is updated. This difference in motivation is reflected in the difference in approaches. Their proposed approaches generally require task-specific data and paired prompts, or require training a prompt transformer using $M_t$ as part of the calculation, which incurs costs on par with re-training the prompt from scratch on the target model. In contrast, our methods only require the embeddings of the source and target models, and can be reused across tasks.

%% file: sections/methods.tex
\section{Methods}

All our experiments follow three steps: 1)~Train a \textit{source prompt} $P_s$ using the \textit{source model} $M_s$ for some task $\Tau$. 2)~``Recycle'' the prompt using some function $r$, learned via the correspondence between the models, that transforms the prompt into one that works on the \textit{target model} $M_t$, $r(P_s) \rightarrow P'_t$. This transformation also handles any changes in the prompt size required when moving between models of different sizes. 3)~Evaluate the recycled prompt $P'_t$ with $M_t$ on the held out split of task $\Tau$. $\Eval(M_t, P'_t, \Tau)$.

Initially, we compare recycling to ``re-tuning'', training a new prompt directly on the $M_t$, for head-room analysis. Given that we found a substantial gap, we judge our recycling methods by how consistently they are able to deliver an improvement over the zero-shot baseline. The zero-shot baseline is the performance of the target model on task $\Tau$ without any task-specific training or prompts, $\Eval(M_t, \Tau)$, relying only on knowledge gained during pre-training. 

Due to the large variance in zero-shot performance (see Table~\ref{tab:perf}) across models, we also compare our recycling performance to the baseline of using a random prompt. The components of our random prompts are drawn from a Gaussian distribution with mean of zero and a standard deviation of $16$, $\Eval(M_t, P_r, \Tau)$ where ${P_r}_i \sim \mathcal{N}(\mu=0,\,\sigma=16)$. We selected $16$ from a grid search over $\sigma \in \{4, 8, 16, 32, 64\}$, and found this to be a surprisingly strong baseline. See Appendix~\ref{appx:random-prompts} for additional details and performance. 

\subsection{Recycling Methods}

We propose two methods for recycling prompts, both based on correspondences between the embedding representations of tokens across the two models. Our hypothesis is that a recycler trained to map embeddings from $M_s$ to $M_t$ can also be used to map prompts. This assumes that 1) prompts are similar to token embeddings, as they are fed into the model the same way 2) there are structural similarities in the relationships between tokens in the embedding spaces of different models, and 3) the relationship between prompt representations and token embeddings are similar across models.

\paragraph{Vocab to Vocab Transformations (\texttt{v2v}):} A mapping between the vocabulary embeddings of two models can be learned and subsequently applied to a prompt. Let $V_s$ and $V_t$ represent the vocabulary embeddings of the source and target models $M_s$ and $M_t$. We wish to find a function $f$ such that $$f(V_s) = V_t$$ and then estimate the target prompt: $$P_t' = f(P_s)$$ 

\texttt{v2v-lin}: In this method, we parameterize $f$ as a linear projection and use least squares to solve for a matrix $Y$ such that $Y V_s = V_t$. We then estimate $P_t' = Y P_s$.

\texttt{v2v-nn}: In this method, we parameterize $f$ with a small neural network, mapping the source embedding of size $E_s$ to the target embedding of size $E_t$, using a hidden dimension of $4 * E_t$ and ReLU activations \cite{fukushimaCognitronSelfOrganizingMultilayered1975,nairRectifiedLinearUnits2010}.

\paragraph{Linear Combination (\texttt{lin-comb}):} 
Another approach is to represent the source prompt $P_s$ as a linear combination of vocabulary embeddings: $$V_{s} X = P_s$$ Once we solve for $X$, we use the same linear combination on the target embedding vectors, for the corresponding tokens, to generate the estimated target prompt: $$P_t' = V_t X$$

Additional details about our recycling methods can be found in Appendix~\ref{appx:recycling-train} 
and implementations have been open-sourced\footnote{\url{https://github.com/google-research/prompt-tuning/tree/main/prompt_tuning/recycling}}.

\subsection{Models}

All models we use are based on T5 1.1 lm100k, a version from T5 1.1 trained for an additional 100K steps with a Language Modeling objective from \citet{lesterPowerScaleParameter2021}. We use the ``Base'' and ``Large'' size version of the model for our cross size recycling experiments. Additionally we trained two more copies of T5 1.1 lm100k Base from scratch using T5X \cite{robertsScalingModelsData2022}, Flaxformer, Flax \cite{flax2020github}, and Jax \cite{jax2018github} from different random seeds. Additional details of pre-training can be found in Appendix~\ref{appx:pretrain}.

In the default setting, Prompt Tuning uses autoregressive generation to make predictions. The prompted model is allowed to generate arbitrary text which is then compared to a predefined answer string---a \textit{verbalizer} \cite{schickItNotJust2021}. In this setting, recycled prompts score zero as they tend to output illegal predictions, i.e.~they generate strings that don't match any verbalizers.

Instead of using generation, we evaluate models with rank classification \cite{brownLanguageModelsAre2020,weiFinetunedLanguageModels2022,minMetaICLLearningLearn2022,sanhMultitaskPromptedTraining2022}. The model is used to score each possible verbalizer, conditioned on the example input and the prompt. The highest ranking class is then selected as the model's prediction: $$\argmax_{y\in\mathcal{Y}} \text{Pr}_{M_t}(y|P_t';X)$$ Thus we only care that the correct class is the most probable of the possible verbalizers, not that it is the most probable generation of all. Recycling success in this setting suggests that, while a prompt trained directly for a given model can learn to control its generated outputs, this ability isn't easily transferred across models.

\input{tables/perf}

\subsection{Source Prompt Training}

To explore how the initialization of the source prompt can influence its recyclability we used ``Random'', ``Class Label'' \cite{lesterPowerScaleParameter2021}, and ``SPoT'' \cite{vuSPoTBetterFrozen2022} initialization methods. For SPoT initialization a prompt pre-trained on language modeling is used as a starting point. Our two SPoT settings use prompts trained for 10K and 50K step respectively. Additional details on the SPoT pre-training procedure can be found in Appendix~\ref{appx:spot}.

We also explore using the initialization and training scheme from \citet{khashabiPromptWaywardnessCurious2022}. The prompt is initialized using the embedded representation of string that describes the task, and a regularization loss during training encourages the learned prompt parameters to remain close to that starting point. We refer to this method as ``Wayward'' initialization. The intuition is that since our recycling methods are grounded in the mapping between model vocabularies, recycling may be more effective if we can keep the source prompts on the manifold of the source model's token embeddings. In this setting we use the words from the text prompt as training data for the recycler. While the ``Wayward'' methodology also includes changes to training in addition to initialization, we evaluate it by comparing it with other initialization methods. 

Additionally, we explore recycling source prompts at various points during their training---specifically after $2$, $5$, $10$, and $20$ thousand steps. \citet{vuSPoTBetterFrozen2022} found that prompts trained beyond 10K steps were less useful for predicting task similarity, suggesting that continued training may result in a prompt overfit to the current task and model. Thus, we hypothesise that prompts from earlier in training will be more transferable.

For each initialization method, we train three prompts per source model, and recycle these to each target model. Details about the training of the source prompts can be found in Appendix~\ref{appx:source-prompt-train}. These trained prompts were also used to calculate the ``Re-tune'' headroom analysis in Table~\ref{tab:perf}.

\subsection{Vocabulary Selection}

Rather than use all $32{,}000$ SentencePiece \cite{kudoSentencePieceSimpleLanguage2018} embeddings for $V_s$ or $V_t$ we use a subset. T5 1.1 was pre-trained exclusively on English data from C4 \cite{raffelExploringLimitsTransfer2020} but shares a vocabulary with the original T5 which supplemented unsupervised pre-training with machine translation. Thus the vocabulary contains non-English tokens (German, French, and Romanian) that have likely never been updated during training. We filter non-English tokens by removing tokens with a cld3\footnote{\href{https://github.com/google/cld3}{https://github.com/google/cld3}} confidence less than $0.8$. Some English subword tokens get removed as well, leaving $16{,}779$ embeddings.
The final list of filtered tokens is available in our open-source release.

\citet{Wendlandt18Surprising} and \citet{TACL1202} observe instability in the local neighborhoods of embeddings between training runs, especially for high frequency tokens. Given that SentencePiece tokens are ordered by frequency, we skip the first $1{,}000$ tokens after filtering as they are more likely to be unstable between models. We then use the next $4{,}000$ tokens as the data points to train our recyclers.

\subsection{Datasets}

We investigate recycling using two sentiment analysis datasets, SST2 \cite{socherRecursiveDeepModels2013} and IMDB \cite{maasLearningWordVectors2011}. Early experiments explored using QQP \cite{WinNT} and ReCoRD \cite{zhang2018record} but found that the target model's zero-shot performance must be non-trivial (better than the naive majority-class baseline) for recycling to work. More information about each task as well as the verbalizers used can be found in Appendix~\ref{appx:datasets}.

%% file: tables/perf.tex
\begin{table*}[t!]
    \centering
    \begin{tabular}{l l | r | r r r}
        Target Model & Dataset & Re-Tune & Zero-Shot & Recycle & $\E_{P_r}$  \\
        \hhline{==|=|===}
        Base  & SST2 & 92.3$_{0.3}$ & 59.2$_{6.6}$ & 64.7$_{8.9}$ & 56.3$_{5.2}$ \\
              & IMDB & 94.2$_{0.2}$ & 65.6$_{8.2}$ & 67.1$_{8.3}$ & 62.8$_{4.1}$ \\
        Large & SST2 & 95.5$_{0.3}$ & 75.0         & 69.6$_{7.6}$ & 70.8$_{4.4}$ \\
              & IMDB & 96.1$_{0.1}$ & 77.2         & 80.3$_{3.4}$ & 81.0$_{0.5}$ \\
    \end{tabular}
    \caption{Recycling lags behind the best-case performance of re-tuning directly on the target model, but shows gains at Base size over zero-shot performance and random prompts ($\E_{P_r}$). Recycling from Base $\rightarrow$ Large slightly underperforms random prompts on average, but this is pulled down by a few very low values; Table~\ref{tab:recyclers} shows that in nearly all settings, the majority of recycling runs exceed the expected performance of random prompts. Results are aggregated over all target models of a given size. For Re-Tune and Recycle, we train prompts to $2$K steps using \texttt{class-init} initialization and aggregate over $3$ random seeds and $3$ recycling methods. For $\E_{P_r}$, $100$ random prompts are sampled. The improvement of recycling over random prompts is statistically significant ($p<0.05$). All results are presented as Accuracy$_{\text{StdDev}}$.}
    \label{tab:perf}
\end{table*}

%% file: sections/results.tex
\section{Experimental Results}


There is a large gap between top-line performance, re-tuning a prompt directly on the target model, and a recycled prompt. Table~\ref{tab:perf} shows that even in the strongest recycling settings, there is still a $15$ point gap. However we do see that recycling prompts yields stronger results than using the target model for zero-shot inference or using a random prompt. This shows that recycling between models---including ones with different sizes---is possible but difficult, and increasing the performance of recycled prompts should be a focus of future research. Due to the size of this gap, the rest of our work focus on improving the reliability of prompt recycling methods. Extra details about the Zero-Shot and Random results can be found in Appendices~\ref{appx:zero-shot-baseline} and~\ref{appx:random-prompts} respectively.

\subsection{Source Prompt Initialization}

\input{tables/init}

First, we explore how source prompt initialization effects the reliability of recycling. As reported in Table~\ref{tab:init}, we find that random initialization results in far lower recyclability on SST2 than the other methods. As such, we do not include random initialization results in our analysis going forward.

We also explore using a pre-trained SPoT prompt for source prompt initialization. Table~\ref{tab:init} shows that SPoT initialization was successful on SST2. Given this success, we also explore SPoT initialization for IMDB but it was less successful. Going forward, results using SPoT initialization are included in aggregations of initialization strategies.

\input{tables/wayward}

We explored using Wayward initialization and training. By initializing prompts trained on different models with a shared text string, regularizing the trained prompt toward that initial value, and uses the embedding of those tokens to learn the recycler we hope to create a shared space between to models, making recycling easier. Due to the training vocabulary token selection (selecting only the tokens included in the text prompt), only the \texttt{v2v-nn} recycler works---recycling prompts with other methods tend to have \texttt{NaN} values. We see in Table~\ref{tab:wayward} that Wayward initialization yields slightly more consistent recycling; however, the resulting models are often much weaker compared to successful recycling of prompts trained with class initialization. Given the minimal improvement to robustness, we do not explore Wayward initialization in other settings or include it in further analysis.

\subsection{Recycling Method Settings}

\begin{figure}[t!]
    \centering
    \includegraphics[width=\columnwidth]{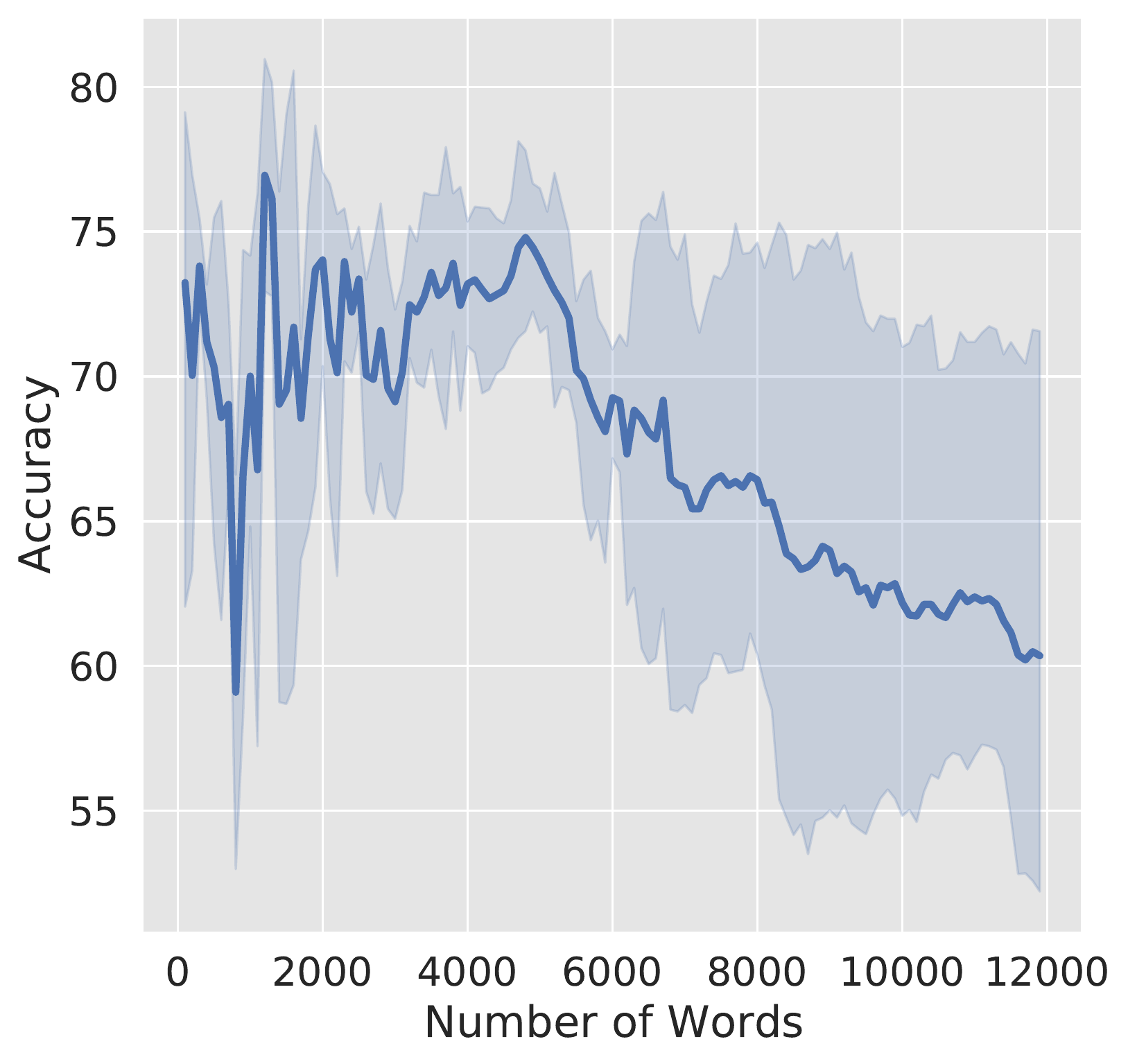}
    \caption{Recycling performance as a function of words used to train the recycler. Each point represents four \texttt{v2v-lin} recycling runs, aggregating over Class Label, SPoT (10K, 50K), and Wayward initialization. The range from $3{,}000$ to $5{,}000$ words delivers relatively high performance and low variance.}
    \label{fig:num-words}
\end{figure}

Figure~\ref{fig:num-words} shows that recycling performance has large variance when using few token embeddings to learn the recycler $r$. It also shows that performance trends downward as more and more tokens are used, possibly due to the relative rarity of these later tokens. We see our chosen number of tokens, $4{,}000$, is well placed between these trends, although $5,000$ may have worked better. We fit the \texttt{v2v-lin} recycler with a variable number of token embeddings ranging from $100$ to $12{,}000$ with a step size of $100$. This recycler is then applied to prompts trained with Class Label, SPoT, or Wayward initialization on the SST2 dataset. Additionally we include results for Class Label initialization using a recycler that skips the first $1{,}000$ tokens.

\begin{figure}[t!]
    \centering
    \includegraphics[width=\columnwidth]{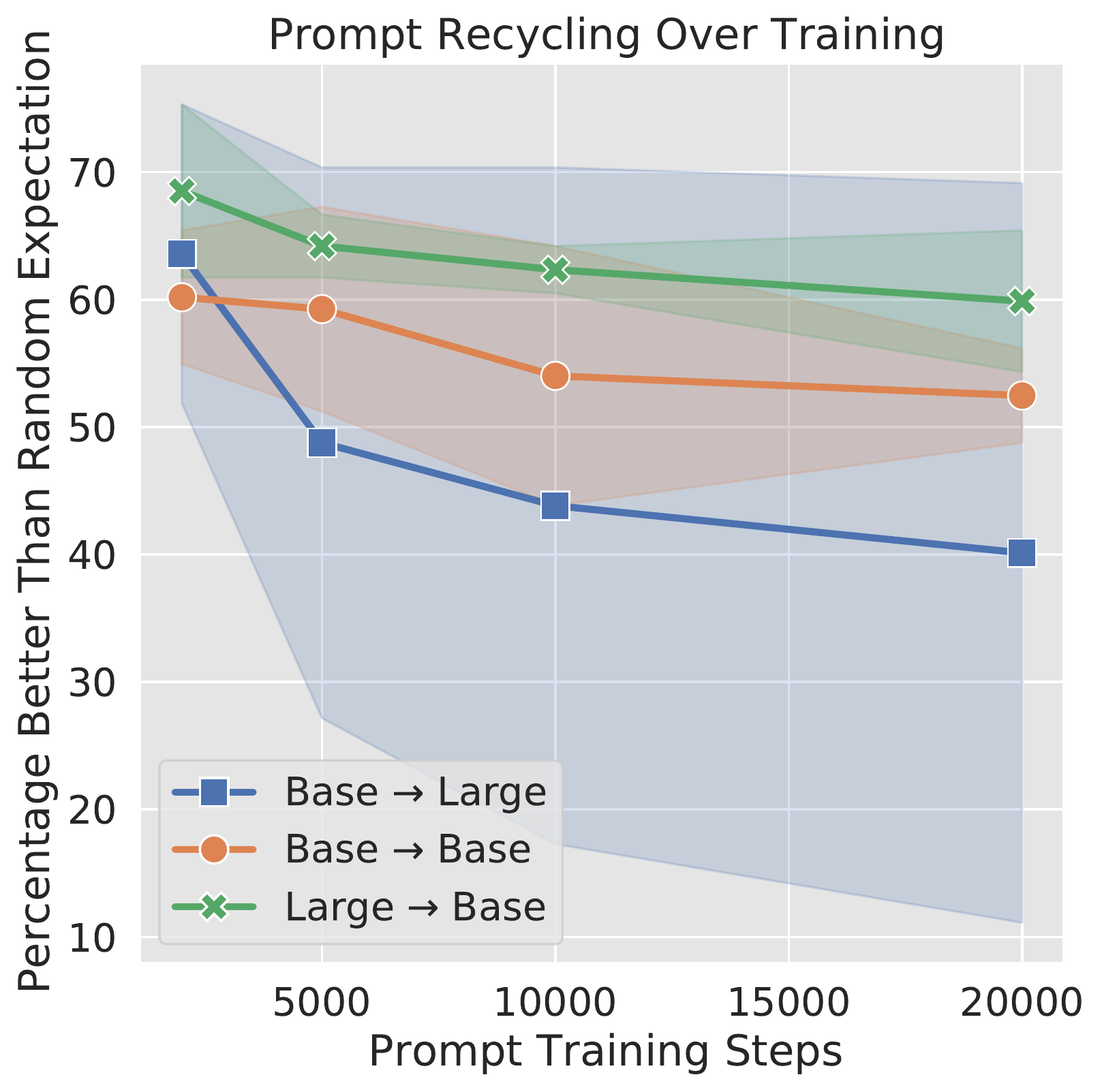}
    \caption{Recyclability drops as the source prompt is trained longer. Results are aggregated over all source and target models of a given size, all possible recycling methods, two initialization methods (\texttt{class-init}, SPoT), and two datasets (SST2, IMDB).}
    \label{fig:over-training}
\end{figure}

As training continues, a prompt begins to see repeated input, output pairs as multiple epochs loop through the dataset multiple times and the prompt becomes more and more specialized for solving this specific task in conjunction with this specific frozen model. Figure~\ref{fig:over-training} shows that recycling prompts later in training results in decreased reliability of recycling. Essentially, the source prompt begins to overfit to the quirks of the specific source model it was trained with. By $2$K steps, trained prompts have converged, they within $2.25\%$ of the maximum performance they will obtain and begin to show increasing loss on the validation set with stagnating accuracy, suggesting the model is getting more and confident in its wrong answers.

\input{tables/recyclers}

Table~\ref{tab:recyclers} shows how different recyclers have very different performance profiles in different settings. We see that \texttt{v2v-nn} is the best method for Base $\rightarrow$ Base recycling while the other methods have poor performance. In the settings where the source and target models are different sizes we see that other methods are stronger, especially for the SST2 dataset.

\subsection{Recycling Across Model Sizes}

\input{tables/upcycle}

We also explore recycling between models of different sizes. Table~\ref{tab:base-to-large} shows that Base $\rightarrow$ Large recycling results in a prompt that is stronger than the expected performance of random prompts over half the time; however, the mean performance of recycling is less than the mean performance of random prompts (see Table~\ref{tab:perf}). This is due to the larger variance of Base $\rightarrow$ Large recycling. In the cases where recycling is stronger it is just a bit stronger, but in the cases it is worse---it is a lot worse. We seen an exaggerated version of this same trend when comparing the Zero-Shot results on SST2.

Table~\ref{tab:base-to-large} shows that Base $\rightarrow$ Large recycling is much more likely to beat the Zero-Shot baseline than Base $\rightarrow$ Base recycler. This appears to be an artifact of the low Zero-Shot performance of the Large models as this does not hold when comparing how often each setting beats the expected random performance, where Base $\rightarrow$ Large is much weaker than IMDB.

\input{tables/downcycle}

\citet{zhongFactualProbingMASK2021} ask if soft prompts perform well because they are better at extracting knowledge already contained within the frozen model or if they are just used as a method to memorize specific information about the training data. Table~\ref{tab:large-to-base} shows that recycling from Large $\rightarrow$ to Base is more robust than Base $\rightarrow$ Base. This suggests that either prompts trained with Large models memorize their task specific information in a way that is easier to transfer to a Base model or that Large prompts leverage information stored in the Large model is a way that translates to knowledge extraction for Base models. This finding is suggestive that better recycling methods maybe able to act similar to distillation methods and transfer knowledge from prompts trained on larger models to smaller ones.

%% file: tables/init.tex
\begin{table}[t!]
    \centering
    \begin{tabular}{l l | r r}
        Dataset & Initialization & \abovezeroshot & \aboverandom \\
        \hhline{==|==}
        SST2 & Random   & 29.6             & 47.22 \\
             & Class    & \underline{50.0} & \underline{66.67} \\
             & SPoT 10K & 50.0             & 58.33 \\
             & SPoT 50K & \textbf{51.9}    & \textbf{69.44} \\
        \hline
        IMDB & Class    & \textbf{72.2}    & \textbf{65.74} \\
             & SPoT 10K & 45.4             & 58.33 \\
             & SPoT 50K & \underline{51.9} & \underline{59.26} \\
    \end{tabular}
    \caption{How often recycled prompts exceed zero-shot (ZS) and random prompt ($\E_{P_r}$) baselines. We aggregate over all source and target models and all recycling methods, for $108$ recycled prompts per row. Prompts trained with Random initialization are far less likely to be successfully recycled. SPoT offers small gains in robustness for SST2, but underperforms on IMDB\@. Bold and underline mark the best and second-best methods.}
    \label{tab:init}
\end{table}

%% file: tables/wayward.tex
\begin{table}[t!]
    \centering
    \small
    \begin{tabular}{l | r r r}
        Initialization & \abovezeroshot & Acc.$_{\text{StdDev.}}$ & \aboverandom \\
        \hhline{=|===}
        Class   &          55.6 & \textbf{69.2$_{7.7}$} & 66.7 \\
        Wayward & \textbf{61.1} &          63.9$_{7.3}$ & 41.2 \\
    \end{tabular}
    \caption{Wayward initialization often beats Zero-Shot but falls short of random and lags on final performance. We aggregate over all source and target models, using the \texttt{v2v-nn} recycler on SST2, giving $36$ recycled prompts per row. Wayward's low accuracy when successful also means it rarely beats random prompts, which score quite well for some models. The difference in Accuracy is statistically significant ($p<0.05$).}
    \label{tab:wayward}
\end{table}

%% file: tables/recyclers.tex
\begin{table*}[t!]
    \centering
    \begin{tabular}{l l l l | r r}
        Source & Target & Recycler & Dataset & \abovezeroshot & \aboverandom \\
        \hhline{====|==}
        Base  & Base  & v2v-nn      & SST2 & \textbf{63.0} & \textbf{64.8} \\
              &       &             & IMDB & \textbf{55.6} & 64.8 \\
              &       & lin-comb    & SST2 & 35.2          & 50.0 \\
              &       &             & IMDB & 38.9          & 64.8 \\
              &       & v2v-lin     & SST2 & 35.2          & 50.0 \\
              &       &             & IMDB & 38.9          & \textbf{66.7} \\
        \cline{2-6}
              & Large & v2v-nn      & SST2 & 55.6          & 70.4 \\
              &       &             & IMDB & 77.8          & 44.4 \\
              &       & lin-comb    & SST2 & 66.7          & \textbf{77.8} \\
              &       &             & IMDB & \textbf{96.3} & \textbf{55.6} \\
              &       & v2v-lin     & SST2 & \textbf{70.4} & 77.8 \\
              &       &             & IMDB & 92.6          & 55.6 \\   
        \hline
        Large & Base  & v2v-nn      & SST2 & 48.2          & \textbf{77.8} \\
              &       &             & IMDB & 37.0          & 51.9 \\
              &       & lin-comb    & SST2 & \textbf{51.9} & 74.1 \\
              &       &             & IMDB & 51.9          & 66.7 \\
              &       & v2v-lin     & SST2 & 48.2          & 74.1 \\
              &       &             & IMDB & \textbf{55.6} & \textbf{66.7} \\       
    \end{tabular}
    \caption{Reliability of three recyclers across various settings. Results are aggregated over source and target models and \texttt{class-init} and SPoT initialization methods, giving $54$ recycled prompts for each Base $\rightarrow$ Base row, and $27$ prompts elsewhere. The \texttt{v2v-nn} recycler is best for Base $\rightarrow$ Base transfer, while the other methods are more reliable when transferring across model sizes. Bold marks the best method for each dataset per block.}
    \label{tab:recyclers}
\end{table*}

%% file: tables/upcycle.tex
\begin{table}[t!]
    \centering
    \begin{tabular}{ l l | r r }
        Dataset & Target & \abovezeroshot & \aboverandom \\
        \hhline{==|==}
        SST2 & Base  & 44.4          & 54.9 \\
             & Large & \textbf{64.2} & \textbf{75.3} \\
        \hline
        IMDB & Base  & 44.4          & \textbf{65.4} \\
             & Large & \textbf{88.9} & 51.8 \\
    \end{tabular}
    \caption{Recycling from Base $\rightarrow$ Large is inconsistent. There are large improvements over Base $\rightarrow$ Base recycling for SST2 but not when comparing to random prompts for IMDB\@. Results are aggregated over all source and target models of a given size, all recycling methods, and \texttt{class-init} and SPoT initialization methods. Results with Base as the target size include $162$ recycled prompts, while results with Large include $81$.}
    \label{tab:base-to-large}
\end{table}

%% file: tables/downcycle.tex
\begin{table}[t!]
    \centering
    \small
    \begin{tabular}{ l l | r r r }
        Dataset & Source & \abovezeroshot & Acc.$_{\text{StdDev.}}$ & \aboverandom \\
        \hhline{==|===}
        SST2 & Base  & 44.4 & 59.5$_{8.8}$ & 54.9 \\
             & Large & 49.4 & \textbf{61.6$_{5.4}$} & 75.3 \\
        \hline
        IMDB & Base  & 44.4 & 69.3$_{5.3}$          & 65.4 \\
             & Large & 48.2 & 69.8$_{6.4}$          & 61.7 \\
    \end{tabular}
    \caption{Recycling from Large $\rightarrow$ Base is more reliable than Base $\rightarrow$ Base recycling, especially for SST2\@. Results are aggregated over all source and target models of a given size, all recycling methods, and \texttt{class-init} and SPoT initialization methods. Results with Base as the source size include $162$ recycled prompts while results with Large include $81$. Bold results show where the gain from Large recycling is statistically significant $(p<0.05)$.}
    \label{tab:large-to-base}
\end{table}

%% file: sections/conclusion.tex
\section{Conclusion}

We have demonstrated that recycling prompts between different models is possible, but difficult. Learned transformation based on structural similarities between the models can be applied to the prompt to create a new prompt, tailored to the target model, that infuses extra information useful for some task beyond the knowledge built into the target model. This manifests as recycled prompts having stronger performance than the target model's zero-shot performance. Additionally, prompts in the restricted area of the possible prompt space dictated by the transformation of a source prompt tend to have stronger performance than randomly sampled prompts. However, recycling is difficult as the final performance of recycled prompts is still far below Re-tuned prompts, trained in conjunction with target model.

We have proposed three different recycling methods and found that different recyclers work better when applied to different model and source prompt combinations. We found Base $\rightarrow$ Large recycling is unstable and that Large $\rightarrow$ Base recycling is more robust and produces stronger prompts than Base $\rightarrow$ Base recycling, and that recyclability tends to decrease the more a prompt is trained.

These successes, and even more the failures with respect to pure performance, demonstrate that prompt recycling, and the idea of correspondences between similar models that it is based on, is an exciting research direction where many improvements can still be made.

%% file: sections/acknowledgements.tex
\section*{Acknowledgements}
We thank Rami Al-Rfou for their feedback on this work.

%% file: sections/appendix.tex
\section{Appendix}

\subsection{Pre-training Hyperparameters}
\label{appx:pretrain}

Two new version of T5 1.1 Base lm100k were trained using the T5X framework. They were initially trained on C4 using the Span Corruption \cite{raffelExploringLimitsTransfer2020} objective with an average span size of $3$ and $15\%$ of tokens removed as part of a span on average. Training was done for $1$ million steps using a batch size of $256$. Input examples were trimmed to $512$ tokens while outputs where trimmed to $114$. All optimization parameters use the defaults from T5X. Afterwards, the model are trained a further $100$K steps using the Language Model objective. Here a batch size of $128$ is used and inputs and targets are trimmed to $1024$ and $128$ respectively. All pretraining was done on 64 TPUv3s.

\subsection{SPoT Pre-training Hyperparameters}
\label{appx:spot}

SPoT initialization vectors were pre-trained on the Language Modeling Objective used to adapt T5 1.1 models in \citet{lesterPowerScaleParameter2021}. Sequences of length $512$ where randomly divided into input and target sequences with a maximum length of $128$. The model is fed the input and must predict the target. Token-level cross-entropy loss is used to calculate gradients and updates are only applied to the prompt parameters. The Adafactor \cite{shazeerAdafactorAdaptiveLearning2018} optimizer is used with hyperparameters matching \citet{lesterPowerScaleParameter2021} (constant learning rate of $0.3$, weight decay of $1e^{-5}$, and parameter scaling turned off). The SPoT prompt is pre-trained for 50K steps and the prompt from step 10K and 50K are used in our experiments.

A single SPoT prompt was trained for each frozen model. The SPoT prompt has a length of $100$ is was initialized using tokens embeddings sampled from the first $5,000$ tokens in the model's vocabulary. SPoT prompts for the Base sized models were trained on 8 TPUv2s while Large sized models used 16 TPUv3s.

\subsection{Recycling Method Training}
\label{appx:recycling-train}

The \texttt{v2v-nn} recycler was trained with JAX and Flax using the Adam \cite{kingmaAdamMethodStochastic2015} optimizer with a batch size of $50$ and a learning rate of $3e^{-4}$ for $25$ epochs.

The \texttt{v2v-lin} recycler was fit using \texttt{tf.linalg.lstsq} \cite{tensorflow2015-whitepaper}. This was also used to finding the linear combination of embeddings that approximate the prompt in the \texttt{lin-comb} recycling method.

\subsection{Source Prompt Training Hyperparameters}
\label{appx:source-prompt-train}

All source prompts are $100$ tokens long and were trained using the best hyperparameters from \citet{lesterPowerScaleParameter2021} (constant learning rate of $0.3$, weight decay of $1e^{-5}$, and parameter scaling turned off). Prompts where trained for 20K steps with a batch size of $128$.
Input examples were trimmed to $512$ tokens and outputs to $16$ (note, all verbalizers were shorter than this limit).

Three source prompts were trained for each frozen model initialization strategy pair. Differences between runs due to the random seed include the initial prompt value and the order training examples.
Source prompts for the Base sized models were trained on 8 TPUv2s while Large sized models used 16 TPUv3s.

The different initialization strategies used the following hyperparameters:
\paragraph{Random Initialization:} The prompt is initialized from a uniform random distribution between $-0.5$ and $0.5$.

\paragraph{Class Initialization:} The prompt is initialized from the embeddings of the possible class verbalizers (where the embeddings for subword tokens are averaged together in the case of a verbalizer being tokenized into multiple SentencePieces). Additional prompt tokens are initialized with token embeddings sampled from the first $5,000$ tokens in the model's vocabulary.

\paragraph{SPoT 10K/50K:} The prompt is initialized with a SPoT prompt after 10K or 50K steps of SPoT pre-training. See Appendix~\ref{appx:spot} for details on the SPoT pre-training procedure.

\paragraph{Wayward:} The prompt is initialized with the SentencePiece embeddings of the following string: ``\textit{Classify this movie review based on its sentiment . Use 2 classes . One positive ( for reviews that paint the movie in a favorable light ) and one negative ( for reviews that make you not want to see the movie or think it will be bad ) . Use the string ` positive ` for the positive class , the good / great movies , and use the string ` negative ` for the negative class , the bad movies .}'' Spaces where added around punctuation to match the text pre-processing used for SST2. This string is tokenized into $100$ tokens by the T5 vocabulary. During training, the prompt parameters were regularized towards this initial representation with a squared $L_2$ distance loss term, normalized by the prompt length.
$$
\mathcal{L}_{\text{wayward}} = \frac{\| P_s - \text{Embed}(\text{prompt}) \|^{2}_{2}}{L}
$$
This loss term is then scaled by the parameter $\gamma=0.01$.

\subsection{Random Prompts}
\label{appx:random-prompts}

\begin{table}[t!]
    \centering
    \begin{tabular}{l | r r}
        Model & SST2 & IMDB \\
        \hhline{=|==}
        Base   & 52.4$_{2.7}$ & 64.7$_{4.6}$ \\
        Base 1 & 58.6$_{4.7}$ & 64.9$_{1.6}$ \\
        Base 2 & 58.4$_{5.2}$ & 59.3$_{2.4}$ \\
        Large  & 71.2$_{4.5}$ & 81.0$_{0.5}$ \\
    \end{tabular}
    \caption{Random Prompt performance of various frozen models. Base 1 and Base 2 are the versions of T5 1.1 lm100k Base that were trained in T5X while all other models were originally trained in MeshTF and converted to the T5X format. Results calculated from draws of $100$ random prompts and presented as Acc.$_{\text{StdDev.}}$.}
    \label{tab:random-perf}
\end{table}

Random Prompt experiments involved drawing a random prompt, $P_r$ and using that random prompt with the target model for task evaluation, $\Eval(M_t, P_r, \Tau)$. The elements of the random prompt are draw from a Gaussian distribution with a mean of $\mu=0$ and a standard deviation of $\sigma=16$, thus ${P_r}_i \sim \mathcal{N}(mu=0,\, \sigma=16)$.

The standard deviation of $\sigma=16$ was selected via a grid search over $\sigma \in \{4, 8, 16, 32, 64\}$. Analysis of trained prompts have shown that the prompt parameters tend to have large values and norms, therefore we made sure to include large values in our grid search.

The expected performance of Random prompts show how easy it is to find a value the results in strong downstream performance. If the recycled prompt is better than random prompts it demonstrates that recycling from source prompts find an area of prompt space the produces stronger results and recycling was able to transfer useful information. Table~\ref{tab:random-perf} shows the average performance of random prompts for each model.

\subsection{Dataset Details}
\label{appx:datasets}

All datasets used are the versions distributed via TensorFlow Datasets \cite{TFDS}. The datasets used either do not include a validation split or have non-public labels for their test split. In the latter case we use the public validation split as the test split and in both cases we use the last $N$ examples from the training split for validation. See Table~\ref{tab:datasets} for dataset details.

\begin{table}[t]
    \centering
    \small
    \begin{tabular}{l l| r r r }
        Dataset & Version & Train & Valid & Test   \\
        \hhline{==|===}
        SST2   & 2.0.0 & 66{,}349 & 1{,}000 & 872 \\
        IMDB   & 1.0.0 & 24{,}000 & 1{,}000 & 25{,}000 \\
        QQP    & 2.0.0 & 353{,}846 & 10{,}000 & 40{,}430 \\
        ReCoRD & 1.0.2 & 90{,}730 & 10{,}000 & 10{,}000 \\
    \end{tabular}
    \caption{The TensorFlow Datasets version number and train/validation/test split sizes for each dataset used. Our validation sets where created by taking the last $N$ examples from the train split. TFDS maintains class label distributions in each sample when sliced like this.}
    \label{tab:datasets}
\end{table}

SST2\footnote{\url{https://www.tensorflow.org/datasets/catalog/glue##gluesst2}} is a sentiment analysis dataset built with movie reviews from \href{https://www.rottentomatoes.com/}{rottentomatoes.com}. IMDB\footnote{\url{https://www.tensorflow.org/datasets/catalog/imdb_reviews}} is also a sentiment analysis dataset but the reviews come from \href{https://www.imdb.com/}{imdb.com}. QQP\footnote{\url{https://www.tensorflow.org/datasets/catalog/glue##glueqqp}} is a duplicate question detection dataset built from questions ask on \href{https://www.quora.com/}{quora.com}. ReCoRD\footnote{\url{https://www.tensorflow.org/datasets/catalog/super_glue##super_gluerecord}} asks a question about a provided news paragraph. The answer is an entity from a provided list. A single real-world entity may appear in the list multiple times as different surface forms. If this is the correct entity, all surface forms are considered correct.

As our models generate strings the correspond to a class (instead of a distribution of scores over possible classes), a verbalizer is chosen to represent each class. See Table~\ref{tab:verb-sst2}--\ref{tab:verb-record} for details of the chosen verbalizers and baseline model performance. 

\begin{table}[t]
    \centering
    \begin{tabular}{l|r}
        Verbalizer & Single-Class Accuracy \\
        \hhline{=|=}
        \texttt{positive} & 50.92 \\
        \texttt{negative} & 49.08 \\
    \end{tabular}
    \caption{Verbalizers used for SST2 and the evaluation accuracy if a model were to only select that label.}
    \label{tab:verb-sst2}
\end{table}

\begin{table}[t]
    \centering
    \begin{tabular}{l|r}
        Verbalizer & Single-Class Accuracy  \\
        \hhline{=|=}
        \texttt{positive} & 50.0 \\
        \texttt{negative} & 50.0 \\
    \end{tabular}
    \caption{Verbalizers used for IMDB and the evaluation accuracy if a model were to only select that label.}
    \label{tab:verb-imdb}
\end{table}

\begin{table}[t]
    \centering
    \begin{tabular}{l|r}
        Verbalizer & Single Class Accuracy \\
        \hhline{=|=}
        \texttt{duplicate} & 36.82 \\
        \texttt{unique}    & 63.18 \\
    \end{tabular}
    \caption{Verbalizers used for QQP and the evaluation accuracy if a model were to only select that label. The default verbalizer for the majority class in the T5 codebase is \texttt{not\_duplicate}. We found that when using verbalizer with so many rare SentencePiece tokens caused issues manifesting as very low zero-shot performance, therefore; we use \texttt{unique} as it is a more common word and it is tokenized into the same number of SentencePieces as \texttt{duplicate}.}
    \label{tab:verb-qqp}
\end{table}

\begin{table}[t]
    \centering
    \begin{tabular}{l|r}
        Verbalizer & Random Guess Accuracy \\
        \hhline{=|=}
        Entity Name & 14.72 \\
    \end{tabular}
    \caption{Evaluation accuracy if a model makes random guesses. ReCoRD doesn't have a single set of possible verbalizers. Instead, a list of entities is provided for each example which are used as possible classes for that example. Above is the expected performance if a model were to randomly guess from that list.}
    \label{tab:verb-record}
\end{table}

\subsection{Zero-Shot Performance}
\label{appx:zero-shot-baseline}

\begin{table}[t!]
    \centering
    \begin{tabular}{l | r r r r}
        Model & SST2 & IMDB & QQP & ReCoRD \\
        \hhline{=|====}
        Base   & 66.5 & 74.0 & 62.3 & 21.5 \\
        Base 1 & 53.9 & 65.3 & 63.2 & 22.8 \\
        Base 2 & 57.1 & 57.6 & 63.2 & 22.8 \\
        Large  & 75.0 & 77.2 &  N/A &  N/A \\
    \end{tabular}
    \caption{Zero-Shot performance of various frozen models. Base 1 and Base 2 are the versions of T5 1.1 lm100k Base that were trained in T5X while all other models were originally trained in MeshTF and converted to the T5X format.}
    \label{tab:zero-shot}
\end{table}

\begin{table}[t!]
    \centering
    \begin{tabular}{l | r r r r}
        Model & SST2 & IMDB & QQP & ReCoRD  \\
        \hhline{=|====}
        Base  & 59.2$_{6.6}$ & 65.6$_{8.2}$ & 62.2$_{1.0}$ & 22.4$_{0.7}$ \\
        Large & 75.0         & 77.2         &          N/A & N/A
    \end{tabular}
    \caption{Zero-Shot performance of various frozen model sizes, aggregated over the multiple versions of the Base model.}
    \label{tab:zero-shot-agg}
\end{table}

Throughout this work, we compared a recycled prompt to the zero-shot performance of the target model. If the recycled prompt performs better than zero-shot, it means that recycling was able to transfer information encoded in the source prompt into the target prompt. Otherwise we would not expect to score higher than the information already target model allows. Table~\ref{tab:zero-shot} shows the zero-shot performance of our models on various datasets and Table~\ref{tab:zero-shot-agg} shows zero-shot performance aggregated over model size.

We see non-trivial zero-shot performance on SST2 and IMDB, the datasets where recycling was the most successful. This suggests that the frozen target model must have some baseline capability in solving a task if recycling is expected to work.

It appears that these models have some ability to solve ReCoRD in a zero-shot manner, but this is actually due to bias in the dataset. In ReCoRD, shorter entities are slightly more likely to be correct. If length normalization is used during rank classification (removing the models bias towards shorter outputs) zero-shot performance falls to the random guess baseline of $14.72$.

\subsection{Composability of Prompts and Generative Inference}

In an effort to remove the requirement of rank classification, we explored composing a recycled prompt with prompt that already knows how to control the output of the target model. Ideally, the recycled prompt would include information on how to decides which verbalizer represents the correct class given the current input and the control prompt would be used make the target model output the actual text of the verbalizer.

First a prompt was trained with $M_t$ on a modified version of the c4 dataset. The input text was the same as the Language Modeling objective but the target was one of the possible verbalizers for SST2 (\texttt{positive} or \texttt{negative}) uniformly sampled. Thus this prompt, $P_v$ is trained to always output one of the verbalizers, regardless of the input. Only $500$ steps of training was required to learn the verbalizer distribution. Then a source prompt $P_s$ was recycled to $M_t$ yielding $P'_t$. This recycled prompt was then composed with $P_v$, through concatenation or broadcasted addition, ($P'_t \circ P_v)$. Finally this prompt is used in conjunction with the target model on the task, $\Eval(M_t, P'_t \circ P_v, \Tau)$, using autoregressive generation instead of rank classification.

Composing $P_v$ with $P'_t$ did change the models output from all illegal predictions (often just ``.'', resulting in an Accuracy of zero) when using just $P'_t$ to valid verbalizers when using the composition. Without an explicit method to ensure the prompts were compatible, the composite prompt tended to output only a single class. However, when the composite prompt did output the other class, it was always correct. This suggests that it may be possible to design future prompt training and recycling methods where parameters that encode task information and model output control are explicitly separated and later combined during recycling.

\subsection{Significance Testing}

All tests for statistical significance uses the Welch's t-test \cite{welchTTest1947} as implemented in SciPy \cite{2020SciPy-NMeth} using the \texttt{scipy.stats.ttest\_ind\_from\_stats} function. We use $p<0.05$ as the threshold for statistical significance.

\subsection{Graphs}

All graphs where made with Seaborn \cite{Waskom2021} and Matplolib \cite{Hunter:2007} where bands represent a confidence interval of 95.